# Algorithmic transparency and manipulation

Michael Klenk

TU Delft

m.b.o.t.klenk@tudelft.nl

**Abstract:** A series of recent papers raises worries about the manipulative potential of algorithmic transparency (to wit, making visible the factors that influence an algorithm's output). But while the concern is apt and relevant, it is based on a fraught understanding of manipulation. Therefore, this paper draws attention to the 'indifference view' of manipulation, which explains better than the 'vulnerability view' why algorithmic transparency has manipulative potential. The paper also raises pertinent research questions for future studies of manipulation in the context of algorithmic transparency.

***

## 1 Introduction

Algorithmic transparency is often considered an essential goal in the responsible design and deployment of AI (Winfield et al. 2021; Felzmann et al. 2020). Relevant information should be disclosed to allow interested parties to monitor, check, criticise, or intervene in decisions by the algorithm (Diakopoulos & Koliska, 2016, 3). But algorithmic transparency may have an overlooked dark side, too. Recently, Wang (2022, 2023) and Franke (2022) discussed a new, critical perspective on the link between transparency and manipulation. In short, disclosing information about an algorithm may normalise behaviour that benefits the operators of the algorithm, which can constitute manipulation in the context of power disparities insofar as it exploits people's vulnerability. Thus, "we need to worry about algorithmic transparency as manipulation" (Wang 2023, p. 1). If this critical perspective is apt, the debate about algorithmic transparency has a severe omission that needs to be addressed.

Drawing attention to the manipulative potential of algorithmic transparency is an important and fruitful project. But, so far, it is based on an inadequate understanding of manipulation. Wang (2022, 2023) relies on the view that manipulation exploits vulnerabilities (which I will call *the vulnerability view* for short) to demonstrate the manipulative potential of algorithmic transparency. The vulnerability view is fraught with issues. For one, the vulnerability view fails to support the conclusion that algorithmic transparency amounts to manipulation because it is unclear how, if at all, algorithmic transparency exploits vulnerabilities.

Moreover, the vulnerability view has independent problems on its own. Exploiting vulnerabilities is neither a sufficient nor a necessary criterion for manipulation. Thus the vulnerability view is ill-equipped to support sound conclusions about categorising influences as manipulation, which hampers the assessment of algorithmic transparency's manipulative potential.

Therefore, another model of manipulation is needed to explore further the critical perspective on algorithmic transparency. This paper aims to contribute to this new research angle and strengthen the critical perspective on algorithmic transparency that Wang (2022, 2023) championed. To do so, I show that the risk concerning algorithmic transparency's manipulative potential can be understood with the *indifference view* of manipulation (Klenk 2020, 2021b). The indifference view explains better why algorithmic transparency has manipulative potential. In short, when transparency is used without concern for, or indifferently to, "revealing reasons to users" (Klenk 2021b, p. 101), it quickly degenerates into manipulation, as explained by the indifference view. The indifference view enjoys independent support and exactly highlights what may go wrong with algorithmic transparency: it may be used without concern to reveal reasons to users, and, for that reason, it is manipulative.

The paper should interest scholars of algorithmic transparency and, more generally, anyone interested in manipulation. The aim is modest insofar as I suggest a new direction for the important discussion about algorithmic transparency's manipulative potential. Several questions about how, exactly, to fill in that new perspective will remain open, as I point out below. The paper should nevertheless provide a fruitful starting point for future discussion.

Apart from lessons about algorithmic transparency's link to manipulation, the paper draws a general lesson about the study of manipulation. The claim that algorithmic transparency is manipulation requires a clear view of what manipulation is to justify and defend that claim ('it is manipulation because….'). When the underlying account of manipulation is misleading, we might be led in false directions, e.g., by searching in vain for features wrongly associated with manipulation. By attempting to make explicit what manipulation is and how a given phenomenon – in this case, algorithmic transparency – satisfies the relevant criteria, we can make progress in understanding the many manifestations of manipulation beyond shaky allegations and conjectures.

Section 2 re-constructs and clarifies Wang's argument. Section 3 critically evaluates the argument, and section 4 introduces the superior indifference view that salvages Wang's argument. I conclude in section 6 and suggest questions for further research.

## 2 Wang on transparency and manipulation

In this section, I offer an interpretation of Wang (2022)'s argument about the link between algorithmic transparency and manipulation (section 2.1 and 2.2) and discuss two clarifications (2.3), which serves as a basis for the critical discussion later.[1]

### 2.1  Situating Wang's argument

The predominant view in scholarship and regulatory debates about algorithms stresses that algorithmic transparency is crucial to algorithmic systems' responsible design and use(e.g., Winfield et al. 2021).

But there have also been critical voices (e.g., Bannister and Connolly 2011). For instance, it is recognised that transparency is not merely a neutral transmission of information but a social process linked to power dynamics (cf. Ananny and Crawford 2018).

In particular, as Wang (2022) notes, others have already made tentative suggestions about the link between transparency and manipulation (Kossow et al. 2021; Wachter et al. 2018). These early discussions, however, remained on a rather general level. Effectively, they suggested that corporate interests behind algorithmic transparency may corrupt its otherwise laudable goals (as a kind of ethics washing). What was lacking was a more detailed perspective on how, exactly, manipulation is related to or caused by algorithmic transparency itself.

---

[1] I focus on the argument in Wang (2022) because it is the most detailed and extensive regarding the conceptual claim that algorithmic transparency can constitute manipulation. Wang (2023) and Franke (2022) seem to agree about the conceptual claim and offer different perspectives on the ethical question of whether and why such manipulation is morally problematic. I will discuss their contributions insofar as they bear on Wang (2022) claims about the ethics of manipulation.

Wang's main contribution, and the subsequent discussion in Wang (2023) and Franke (2022), take these tentative suggestions further. It is a novel contribution insofar as it may show in some detail how algorithmic transparency *itself* can constitute or lead to manipulation (Wang 2022, p. 2).

## 2.2 Re-constructing Wang's argument

Wang outlines a specific process – the objectification of norms that results from algorithmic transparency – and argues that this process qualifies as manipulation. In his argument, Wang relies on the view of manipulation developed by Susser et al. (2018), which I call the *vulnerability* view for short (see also Susser et al. 2019). Let us now look at the individual steps in Wang's argument.

The FICO algorithm serves Wang as an illustrative example of algorithmic transparency's manipulative potential. Over 90% of lenders in the US use the FICO algorithm to determine the creditworthiness of individuals. Arguably, the FICO algorithm is transparent in the informational sense because there is publicly available information regarding (1) the categories of data collected, (2) the sources and techniques used to acquire that data, and (3) the specific data points that a tool uses for scoring (Hurley and Adebayo 2016, 204, 213). Because of this transparency, argues Wang, there is a risk of manipulation.

First, Wang defends the empirical premise that algorithmic transparency leads to the 'objectification' of norms (Wang 2022, p. 13).[2] For example, the FICO

---

[2] There are several critical questions about the norm-objectification premise that I will set aside for the purposes of this paper. First, it is unclear whether Wang wants to show or has shown that the system is indeed *intended* for that purpose, or whether it is really a side effect. I will set

algorithm rewards punctual payment of bills, and this information suggests that paying on time is a norm, effectively "disciplining [consumers] according to some expected norms of being responsible credit consumers "(Wang 2022, p. 12). Individuals gradually come to think of these norms "as natural and necessary" and accept them without critical analysis: the norms become "objectified" (Wang 2022, p. 12). They come to think of the system in a particular way (e.g., as objective and value-neutral in the case of the FICO algorithm), and their minds are "reframed" so that their "thinking of other possibilities" is "constrained" (Wang 2022, p. 15). For example, they fail to see the FICO algorithm as an "arbitrary," "discriminatory," or "unfair" system because they are more likely "to only focus on the scientific and objective narrative of its algorithm, ignoring other alternative narratives" (Wang 2022, pp. 16–17).[3]

Second, Wang suggests that the objectification of norms constitutes manipulation by drawing on the vulnerability view. Wang adopts a version of the vulnerability view from Susser et al. (2018), who argue that manipulation "exploit[s] the manipulee's cognitive (or affective) weaknesses and vulnerabilities in order to steer his or her decision-making process towards the manipulator's ends" (cf. Wang 2022, 2, 18).[4] Wang does not explicitly say how norm objectification

---

that issue to the side in this paper. Moreover, there are several open empirical questions about Wang's norm-objectification premise. I set them aside in this paper to focus on the manipulation aspect of his argument.

[3] Wang suggests that algorithmic transparency "opens the black box" so that people know what the rules are and can actively try to conform to them", cf. Wang (2022, p. 13). Consumers can indirectly derive, and are directly told, about an „ideal model" of someone that the algorithm would rate highly. In various ways, people may be influenced to conform to the model. Given the rewards and punishments associated with creditworthiness, "consumers as rational individuals will try to better their position" behaving in ways "to their advantage," cf. Wang (2022, p. 13). For example, upon learning that "payment history" will be considered in FICO's algorithm, individuals would tend to make prompt repayment to improve their credit scores.

[4] Wang does not fully adopt Susser et al's (2019). As I discuss in more detail in section 3, Susser et al (2019) defend *covert influence* as a necessary criterion for manipulation, whereas Wang

exploits vulnerabilities, nor does he define what 'vulnerabilities' are in general.[5] He seems to suggest, however, that norm-objectification is a way to 'exploit the manipulatee's weaknesses and vulnerabilities' to benefit the manipulator in contexts of "asymmetrical power relations," such as commercial and political settings (Wang 2022, 3, 17).[6] For example, the FICO algorithm is "a commercial tool for lenders to make profits" (Wang 2022, p. 19); transparency about the algorithm can lead to "disciplined" individuals that follow the newly objectified norms about behaving in line with good credit-rating scores so that they can be charged with higher interest rates at lower risk of default, which benefits the lender (Wang 2022, p. 19). This, argues Wang, constitutes manipulation on the vulnerability view.[7]

---

often highlights that – to the contrary – manipulation can take place non-covertly (e.g. Wang 2022, 69). Thanks to an anonymous referee for prompting me to clarify this point.

[5] Unlike Wang, Susser et al. (2018, p. 40) distinguish between general (shared by "all human beings in virtue of their embodied condition") and "situated, socially constructed, or contingent vulnerabilities." They further distinguish the latter into structural vulnerabilities, which derive from membership in groups with differential levels of advantage (e.g. being poor, or of a certain gender), and individual vulnerabilities, which are irrespective of group membership and derive, e.g., from one's personal history or habits. Susser et al. (2018, p. 41) write that contingent vulnerabilities are not "monolithic" and that various overlaps and combinations of vulnerabilities can pertain to any one person. This makes it understandable why they characterise online manipulation, a type of influence that can be highly personalised and targeted, in light of vulnerabilities which, on their view, are also highly personalised and non-monolithic.

[6] See also Wang (2023, p. 2).

[7] Wang (2023), responding to criticism by Franke (2022) of the norm objectification premise, notes that manipulation may also occur by other means. For example, he notes that companies may also manipulate people's behaviour "directly" by changing people's choice architecture, rather than through the process of norm-objectification, see Wang (2023, p. 2). It seems that this interpretation is clearly true: there are many other ways in which people can be manipulated, apart from some process of norm-objectification, e.g. by altering people's options. But that interpretation is not relevant for the claim about transparency as manipulation. The relevant, but doubtful, interpretation is that transparency itself has some role to play in these other ways of manipulation. That interpretation is doubtful because it is completely unclear what these 'other ways' might be in which transparency can manipulate without exploiting norm-objectification. Thanks to an anonymous referee for stressing this point. It seems that the relevant interpretation supports the reconstruction of Wang's argument offered above: norm-objectification is a specific process or way in which vulnerabilities can be exploited. In that sense, the exploration of the link between manipulation and transparency on the indifference view are a charitable contribution to Wang's suggestion that there may be 'other' ways in which algorithmic transparency can be manipulative.

So, according to Wang (2022), algorithmic transparency leads to norm objectification and, in the context of power disparities, operators of algorithms can exploit that vulnerability to steer people toward behaviours that benefit themselves. This, he argues, constitutes manipulation.

## 2.3 Two clarifications

Two clarifications are in order. First, there is an ambiguity in Wang's argument between norm objectification being constitutively or causally linked to the exploitation of weaknesses and vulnerabilities. On the one hand, Wang writes that algorithmic transparency can "lead to" manipulation (Wang 2022, p. 17), which suggests the *causal interpretation* that manipulation can be a result or effect of algorithmic transparency. On the other hand, he also writes that algorithmic transparency *itself* is "potentially manipulative" (Wang 2022, p. 18), thus suggesting a *constitutive interpretation*.

This ambiguity matters for interpreting the connection between the claim that algorithmic transparency causes norm objectification and the claim that this is manipulation. If norm objectification *constitutes* the exploitation of vulnerabilities, then by instigating a process of norm objectification, one is immediately in the business of manipulation. In contrast, if the *causal interpretation* is correct, then norm objectification only *causes*, perhaps contingently, the exploitation of vulnerabilities, and by instigating norm objectification one has not yet automatically committed manipulation. In what follows, I will discuss both interpretations: the constitutive and the causal interpretation of Wang's claims.

Second, Wang does not adequately represent the vulnerability view of manipulation he adopts from Susser et al. (2018).[8] Susser et al. emphasise that "the only necessary condition of manipulation is that the influence is hidden" (Susser et al. 2018, p. 27). Call this the *covertness criterion*. Wang does not mention the covertness criterion in his discussion of the vulnerability view. More so, he might even reject the covertness criterion when he approvingly refers to Estop (2014), writing that "power can operate through transparency to manipulate people—not only through hidden lies but through the transparency of 'truth'" (Wang 2022, p. 5).

In any case, Wang (2022) focuses on processes that exploit someone's weaknesses and vulnerabilities in his discussion of manipulation. But, according to Susser et al. (2018), these are not necessarily criteria by which we can tell whether or not a given influence is manipulation. They are "*the means* through which a hidden influence is imposed" (Susser et al. 2018, p. 27, emphasis added). The means by which something is achieved need not constitute criteria for that thing, and it is important not to confuse criteria and means. In analogy, police activity sometimes involves physical violence e.g., to detain suspects. But physical violence is a means associated with police activity but not a reliable criterion by which we can tell whether or not we are dealing with police activity. Likewise, exploiting vulnerabilities as way in which manipulation often happens need not be a reliable criterion by which we can tell whether we are really dealing with manipulation or some other, perhaps benign, form of influence.

---

[8] E.g. In fn 1 of his paper, Wang (2022) writes that he "follows the understanding of manipulation" given by Susser et al. (2018).

The clarification of the vulnerability view matters because it has implications for assessing Wang's argument. On the one hand, algorithmic transparency must credibly satisfy the criterion of hidden influence to have any chance at qualifying as manipulation, if the vulnerability view is maintained. On the other hand, Wang needs to make it plausible that exploiting vulnerabilities effectively forms a sufficient criterion for manipulation. This seems to be a challenge because, so far, nobody has explicitly defended that criterion: Neither Susser et al. (2018; 2019) nor Wang (2022) explicitly argue that exploiting vulnerabilities is sufficient for manipulation. We will see below how this is a problematic omission for Wang's argument.

## 3  Evaluating Wang's argument about transparency and manipulation

So far, I have re-constructed Wang's argument and clarified two points regarding the vulnerability view. I will now turn to a critical assessment. I argue in this section that Wang's transparency-manipulation argument fails *if* we adopt the vulnerability view of manipulation.

### 3.1   The vulnerability view does not support Wang's conclusion

The vulnerability view does not support the conclusion that algorithmic transparency constitutes or causally leads to manipulation.

Recall that, according to the constitutive interpretation of Wang's argument, norm objectification due to algorithmic transparency *constitutes* the exploitation of vulnerabilities in the context of power disparities and, therefore, amounts to manipulation. This interpretation is not convincing for two reasons.

First, it is implausible to treat norm-objectification *itself* as a vulnerability (and thus 'using' that process cannot count as exploiting a vulnerability).[9] Humans are social animals that follow both descriptive and social norms. Our human capacity for and propensity to follow norms is likely an evolutionary adaptation (Bicchieri 2006; Elster 2015). As part of our general propensity to follow norms, we regularly treat them as objective, which may, in many cases, also be adaptive (e.g. Bowles and Gintis 2013). That something is natural does not mean that it is ethical or legitimate. But treating norm-objectification itself as a vulnerability would leave open why only this and not other universal human traits like pro-sociality constitute vulnerabilities.

Second, it is also implausible to treat norm-objectification as constituting a vulnerability *only* in the context of power disparities. Norm-objectification has certain epistemic costs that arise independently of the context. In the broadest sense, it may lead to an 'unexamined life' which, as Socrates emphatically put, is not worth living (cf. Franke 2022, p. 5).[10] For example, one may believe falsehoods, suffer inadequate understanding, and misinterpret reasons for behaviour. Failing to 'make up your mind' (as Wang puts it) about a situation and an uncritical attitude means that you miss out on these intrinsic and instrumental goods, and these losses arise even if there is no context of power disparity. Therefore, this interpretation does not sit well with Wang's repeated and emphatic emphasis on the relevance of power disparities as an enabling condition for the threat of manipulation by algorithmic transparency.

---

[9] In terms of Susser et al. (2018, p. 40)`s account, norm-objectification may at best be an "ontological" vulnerability, rather than a contingent vulnerability. Their account of manipulation, however, focuses on the latter as the relevant type of vulnerability in the context of manipulation.

[10] Franke (2022) contrasts Socrates' dictum with Whitehead's (1911) emphatic emphasis of the value of automating thought and behaviour in the sense of "extending the number of operations we can perform without thinking about them" (1911, pp. 45–46), cited in Franke 2022. Franke is right to challenge an uncritical adoption of the thought that conscious reflection and deliberation is, per se, valuable. It is beyond the scope of this article to enter into a debate about the respective merits of the positions emphasised emphatically by Socrates and Whitehead. Wang's point about the ability to make up one's mind, and the reference to the importance of revealing reasons adopted by the indifference view (see section 4), can be appreciated at least in the minimal sense that *there are some contexts* when this is valuable (without claiming that this is valuable all the time). Thanks to an anonymous referee for prompting me to clarify this point.

So, Wang's argument fails on a constitutive interpretation because it is not convincing that norm-objectification *exploits vulnerabilities* in the context of power disparities. Norm-objectification is neither a general vulnerability nor dependent on power disparities in whichever wrong-making features it has.

This means that Wang's argument depends on the causal interpretation. According to the causal interpretation, algorithmic transparency leads to norm-objectification, which then *causally contributes to* the exploitation of vulnerabilities in the context of power disparities. To assess this interpretation, recall the case of transparency about the FICO algorithm.[11] Paying bills on time is a relevant criterion for the FICO algorithm, and let us assume the information shared about the algorithm represents this accurately; the (objectified) norms that people follow actually help them to satisfy that criterion. For instance, they end up paying their bills on time, they believe that everyone should do so, and correctly believe that the algorithm judges their behaviour according to the criterion. Transparency in this case is accurate, not misleading, and desirable to many.

The norm-objectification that may follow from transparency in such cases does not amount to the exploitation of vulnerabilities by itself. Of course, the underlying *system* may be unfair and it may be unfair precisely because it reflects and perpetuates power disparities. But that observation is immaterial for Wang's argument, which aims to show that *transparency about a system* rather than the system itself can be manipulation.[12]

---

[11] When transparency means that false or misleading information is communicated about the algorithm, transparency conceivably causes exploitation. Perhaps there will be manipulation as a result. But such a case is obviously irrelevant for Wang's argument to the effect that informationally adequate, genuine transparency can lead to manipulation. This situation must be set aside.

One might think that transparency makes matters worse, aggravating injustices in situations of power disparities. That is, one might think that if a system is rigged against you in some way, then being transparent about the system does not obviously resolve the problem, and indeed it may make it worse if, by a process of norm-objectification, the unfair system remains in place. The claim may be that transparency may lead people to be 'locked in' an unfair system by exploiting their tendency to objectify norms. In the example of the FICO system, people may (erroneously) think of the system as objective and neutral and thus fail to challenge it (as, perhaps, they should).

But it is not the case that transparency makes matters worse. One problem is that this interpretation overlooks the distinction between being harmed ('locked in' the system through objectified norms longer than they would be otherwise) and not being benefitted (being 'released' from the problematic system by de-objectifying norms at a time at which they could be released). On the causal interpretation we are now considering, norm-objectification causally leads to harm. Insofar as harm is understood narrowly as 'less on some relevant dimension than the status quo,' then transparency leads to harm only insofar as it worsens the status quo, i.e., transparency *produces* harm that would not be there were it not for the norm-objectification. But insofar as a problematic, exploitative system is already in place, the status quo is not altered by norm-objectification: matters do not get worse, and transparency does not amount to *additional* harm. The system could be toppled, of course, but that amounts to a potential benefit that is not realised.[13] Plausibly, exploiting vulnerabilities means that someone is harmed, rather than merely not benefitted, and it is quite unclear whether and why anyone is harmed, rather than not benefitted, in that situation by the transparency.

---

[12] Indeed, Wang explicitly contrasts his account with an earlier discussion in Kossow, who suggests that when the dominant structure is dogmatic, it is not only useless to promote transparency, but can re-strengthen the existing power asymmetry, see Kossow et al. (2021).

Another problem is that it is entirely unclear, from an empirical perspective, whether norm-objectification would actually perpetuate the system and what the relevant comparison class is. Unfair systems of power disparity presumably do not rely on transparency to remain in place. And it is hardly conceivable that *not communicating* about an existing system of power disparity at all would be preferable to informational transparency about it. This shows that norm-objectification through transparency itself does not cause the exploitation of vulnerabilities.

Thus, what's problematic is, again, the underlying system and not transparency about it. In situations of power disparity, people may be vulnerable, and the powerful may set up systems that exploit them. But transparency about those systems cannot be considered to lead to additional exploitation of vulnerabilities itself. Therefore, Wang's argument fails, if indeed it depends on the vulnerability view of manipulation.

---

[13] The alternative, *wide* notion of harm would count as harmful anything that does not contribute to an improvement of the status quo. Though I cannot argue for it here, however, that seems to me to be an implausible notion of harm. In any case, the present argument stands independently of that dispute insofar as there is no empirical evidence that non-transparency would lead to benefits, i.e. improvements over the status quo (thus, even if *not* procuring these benefits counts as harm on a wide notion of harm, it is simply empirically unclear whether the benefits would materialise). Thanks to an anonymous referee for prompting me to clarify this point.

## 3.2   The vulnerability view is itself problematic

The vulnerability view is also problematic *independently* of considerations about Wang's argument. This means that even if we could salvage the claim that norm-objectification exploits vulnerabilities, we cannot readily conclude that this is manipulation.

To begin with, the vulnerability view does not provide a sufficiency criterion for manipulation. Such a criterion would be helpful, however, to tell whether or not a given influence is manipulation. Arguably, it would be required if an account of manipulation aspires to approximate something like an explanation of the 'nature' of manipulation. Due to the lack of a sufficiency criterion, the vulnerability view cannot be used to infer, without further argument, that algorithmic transparency amounts to manipulation. The question is thus whether the vulnerability view could give us a sufficient criterion that is plausible.

Proponents of the vulnerability view may suggest that the exploitation of vulnerabilities could do as a sufficiency criterion or, at least, as a reliable sign of manipulation. But this will likely not do.

First, as I noted in section 2.3, we should keep processes or mechanisms of a phenomenon apart from criteria to identify the phenomenon. Second, and more importantly, the vulnerability view itself suggests that the exploitation of vulnerability cannot be a sufficient or reliable criterion that can help us determine whether something is manipulation.

This is because the vulnerability view defends the covertness criterion (or hidden influence, see above) as a *necessary* criterion for manipulation (Susser et al. 2018). But vulnerabilities can also be exploited in an overt, obvious way. For example, a manager can make perfectly clear to their employee that they will face dire professional consequences if they do not abide by the manager's inappropriate wishes. Though this is a case of exploiting vulnerabilities, it is not a case of manipulation according to the vulnerability view itself. So, treating 'exploiting vulnerabilities' as a reliable sign of manipulation, let alone a sufficient criterion, is false by the lights of the vulnerability view itself because the manipulative influence is sometimes in clear sight.

Of course, the covertness criterion has been subject to persuasive challenges in the philosophical literature on manipulation (e.g. Noggle 1996; Barnhill 2014). Next to several counterexamples that challenge the covertness criterion (see especially Barnhill 2014), there are also fundamental moral and conceptual reasons against that criterion. For instance, Klenk (2021b) suggested that the criterion may imply that responsibility for manipulation is shifted toward the victim in problematic ways. After all, an influence counts as manipulation, according to the vulnerability view, only insofar as it remains hidden. If that means that the victim can simply 'undo' the manipulation by being sufficiently aware, countermeasures to manipulation may, inappropriately, focus on educating victims, rather than disciplining manipulators. In light of these challenges, there may be good reason to give up the vulnerability view's core commitment to the necessity of the covertness criterion.

But with its core criterion in trouble, and no sufficiency criterion in sight, it is not clear what the vulnerability view amounts to as a view of manipulation. Most importantly, the vulnerability view does not help demarcate manipulation from non-manipulative influence: we are neither given a sufficient or reliable criterion nor a plausible necessary criterion.

Therefore, there is good reason to question the vulnerability view as an adequate view of manipulation. This means that Wang's argument fails, insofar as it depends on the vulnerability view.

In summary, there is double trouble for Wang's argument. It purports to show that algorithmic transparency leads to norm-objectification (1) *which exploits contingent vulnerabilities*, and (2) *thus counts as manipulation*, according to the vulnerability view. Both points are in serious doubt, as shown in this section. Given the vulnerability view, the argument about transparency's potential manipulation fails.

So far, I re-constructed and criticised Wang's argument about the potential manipulativeness of algorithmic transparency. I concluded that his argument fails if indeed we have to stick to the vulnerability view of manipulation. If that were all there is to it, the critical perspective on algorithmic transparency developed by Wang (2022) would be in trouble.

However, I believe that Wang's intuition about the manipulative potential of algorithmic transparency is on the right track. The problems that I discussed in section 3 originated from the vulnerability view of manipulation, which required us to look for a sense in which algorithmic transparency constitutes or causes the exploitation of vulnerability, which was not to be found.

# 4 The indifference view and manipulation by algorithmic transparency

In this section, I introduce the indifference view of manipulation as a superior alternative to the vulnerability view and show how it salvages Wang's intuition about the manipulative potential of algorithmic transparency.

Wang is correct that algorithmic transparency can amount to manipulation. In short, when algorithmic transparency does not aim to reveal reasons to people but merely aims at achieving a certain effect, such as instigating a particular behaviour or creating a certain impression, when it is, in a slogan, indifferent to reasons (cf. Klenk 2021a), then it degenerates into manipulation. This view, the indifference view, not only has the advantage of salvaging Wang's conclusion (that algorithmic transparency has manipulative potential), but independent and general considerations about the nature and ethics of manipulation also support it.

## 4.1 The indifference view of manipulation

The indifference view of manipulation defines manipulation as an influence that aims to be effective but is not explained by the aim to reveal reasons to the interlocutor (Klenk 2021b).[14]

For example, when a politician uses an image of 'foreign-looking' people in their political ad, and they chose that image *because* it will ignite people's xenophobia and racial hatred and not *because* (implausibly) the image will reveal to people why they have (or lack) reasons to vote for the politician, then the politician is manipulating people (cf. Mills 1995). Similarly, when a recommender system is set to display content that effectively engages people's attention, and it displays that content for that purpose rather than to reveal reasons to users e.g. about whom to vote for, what to buy, or what to believe, then the recommender system is used manipulatively (Klenk 2022, 2020).

---

[14] Ideas pertinent to the indifference view have also been defended by Gorin (2014b), Mills (1995), and Baron (2014). Klenk (2021) uses the term 'carelessness,' whereas Klenk (2022) introduces the more appropriate term 'indifference' to avoid the misleading impression that manipulation is, overall, lazy or not planned out. Indeed, manipulation is often carefully crafted influence in its aim to be effective, but careless or indifferent only to the aim of revealing reasons to others.

The indifference view thus identifies manipulation based on two conditions. First, it only looks at influence that is aimed at a particular goal. In that sense, and in line with most if not all literature on manipulation, the view excludes influence that is purely accidental from counting as manipulation (see Noggle 2018). Second, the indifference view then asks why a particular means of influence was chosen to achieve the relevant goal. Manipulative influence is characterised negatively, in terms of the manipulator's choice of a means of influence that is not being explained by the aim to reveal reasons to the interlocutor. The manipulator is, in that sense, "careless" (Klenk 2021b) or indifferent to, revealing reasons to their victims. Since the account focuses on the grounds for choosing means of influence rather than the goal that is pursued, it is possible that the goal is to benefit the target. This means that the indifference view makes room for paternalistic manipulation, or manipulation that – overall – benefits the target.[15]

---

[15] Franke (2022, p. 4), discussing Wang's example of the FICO algorithm, helpfully points out that, abiding by (objectified) norms can be in the interest of the 'victims' of manipulation. Hence, any account of manipulation used to show how norm-objectification can be manipulative should be compatible with manipulation that benefits the victim. As suggested, the indifference account is compatible with paternalistic manipulation. See Klenk (2021a) for discussion. Accounting for paternalistic manipulation is possible on other theories of manipulation, too. See e.g., Noggle (2020) Thanks to an anonymous referee for prompting me to clarify this point.

The indifference view of manipulation is part of a broad family of views on the nature of manipulation that emphasise the *norm-violating* character of manipulation (Barnhill 2014; Gorin 2014a, 2014b; Noggle 1996, 2020).[16] This perspective suggests that manipulation is a kind of influence that falls short of some ideal. Proposals concerning the nature of the ideal differ. For example, Noggle (1996, 2020) argues that manipulation is an influence intended to make the victim violate some norm of belief, desire, or emotion. Barnhill (2014) provides a broader view, suggesting that manipulation may sometimes be an influence that makes someone behave in non-ideal ways, namely in a way that violates their self-interest. The indifference view takes yet a broader, and indeed quite different, perspective on the ideal in question. Unlike the views of Noggle and Barnhill, for example, the indifference view suggests that the ideal in question concerns the motivation of the manipulator, not the behaviour of the patient. Manipulation occurs when the genesis of the manipulator's influence falls short of an ideal, namely that it is *not* explained by the aim to reveal reasons to the interlocutor. A relevant consequence of this perspective, one that will occupy us further below, is that manipulation need not be the result of nefarious, evil intentions to do wrong. Instead, it can be simply – but perhaps not less problematically – the result of carelessness and indifference.

A relevant question about the indifference view is a potential ambiguity between a strict and a wider reading of the indifference view. On a wider reading, facts about the target play a role in determining whether or not an influence is

---

Noggle (2020)[16] See Noggle (2018) and Klenk and Jongepier (2022) for critical discussion and overviews.

manipulation. On the narrow reading, only the motives of the manipulator count. This is suggested, e.g. by Klenk (2022, p. 112) when he writes that manipulation comes down to "a lack of care [by the manipulator] to reveal reasons to the manipulatee," which says "a lot about the manipulator and next to nothing about the manipulatee." Similarly, Noggle (2018) paraphrases the view as emphasising only the motives of the manipulator, leaving out any reference to what happens 'in' (e.g. whether emotional processing is used) or to the victim (e.g. whether the victim is exploited). In this article, I hew close to existing expositions of the indifference account and thus adopt a narrow reading. An implication is that an attempt to reveal reasons through algorithmic transparency that fails to do so is not manipulative, whereas an attempt to use algorithmic transparency toward some other end that happens – perhaps by chance – to reveal reasons is manipulation. The narrow reading should not obscure, however, that facts about the targets may count toward our *moral evaluation* of manipulation even if they play no role in *defining or conceptualising* manipulation. As such, the vulnerability of people may play a role in our assessment of whether and why manipulation is morally problematic. It is bad enough that manipulators are not properly motivated, as it were. But if their irresponsible influence (contingently) leads to further, negative consequences in light of the vulnerability of people, then that is all the more reason to worry about manipulation. Moreover, it matters how manipulators perceive their targets, since their perception of their targets will influence what it means for them to be motivated to reveal reasons to them.[17]

---

[17] To further illustrate the point, consider a world of omniscient, hyper-rational beings that are not vulnerable at all. Whether or not someone strives to reveal reasons to them or not does not matter at all because they are perfect trackers of reasons. Manipulation on a narrow reading of the indifference view would appear much less of a problem insofar as it will have no discernible

With this sketch of the indifference view on the table, it is helpful to briefly note two relevant contrasts with the vulnerability view. First of all, while the vulnerability view focuses on what actually happens to the patient (are they vulnerable? Are they exploited? What actually goes on in the patient?), the indifference view focuses on the agent (what explains their method of influence?). This agent-focused perspective of the indifference view will help us explain better how and why algorithmic transparency has manipulative potential. Second, even though the indifference view focuses on the agent-perspective, it does not require strongly nefarious intentions such as the intention to 'exploit' the victim or harm them otherwise, but associates manipulation with a characteristic indifference toward the ideal of reasoned discourse. This features make the view well posed to explain manipulation in settings such as a marketplace, where actors are simply out for their own good, and often ruthlessly so, but where it would be misleading to describe them as intentionally out to harm others.[18]

---

consequences on the targets. This does suggest that facts about the potential targets of manipulation – such as their vulnerability – is relevant at least in two ways. First, for our assessment of the importance of manipulation in general and, second, for the moral assessment of a specific instance of manipulation. One can consistently adopt the narrow reading of the indifference view for purposes of defining or conceptualising manipulation and acknowledge the significance of consequences for evaluating manipulation. It is a further question whether the strict reading aligns with intuitions about manipulation. Since it mirrors how, for example, we talk about deception (a deceiver can accidentally make people believe the truth), I take it that the narrow reading enjoys sufficient support; see also Klenk and Jongepier (2022). I thank an anonymous referee for pressing me to clarify this point and for providing a version of the helpful example discussed in this footnote.

[18] Since this is but a sketch of the indifference view (and necessarily so, in view of the aim of the article), relevant questions remain concerning, for example, the precise nature of the ideal to reveal reasons to the interlocutor, and an adequate justification of that ideal (see Noggle (1996) and Hanna (2015) for pertinent discussion about the objectivity of the ideal in question). For the purposes of this article, however, the view is adequately described to explore the implications for the manipulative potential of algorithmic transparency.

## 4.2 Indifferent algorithmic transparency is manipulation

The indifference view explains nicely what might be manipulative about algorithmic transparency. In short, algorithmic transparency may *not* be designed to enhance the decision making capabilities of the users of the algorithm by revealing reasons to them. If that is the case, then algorithmic transparency will be manipulative.

There is some reason to think that at least some instances of algorithmic transparency are manipulative for being indifferent. The operators of an algorithm may be transparent for all sorts of reasons, and the reasons that motivate them in choosing a particular *method* of transparency may not always be to reveal reasons to the users. Instead, they may publish information simply to serve the aim to comply by some regulatory demand, to appear in a certain light and to leave a certain impression on users, to make users behave in a certain way, or simply because a certain functionality that enables transparency is available in a pertinent software library that the developers are using.[19]

Notably, the pertinent point is not the motivation to be transparent in the first place but why particular means or methods of being transparent have been chosen. These two things can come apart.

For example, an organisation may decide to be transparent about their algorithm because they are convinced that it is, ethically, the right thing to do. Still, the organisation faces a question about *how* to achieve algorithmic transparency, that is, what means or methods to employ. They might have the option, for example, of using text to communicate or to record brief instructional videos. There is some evidence that videos enhance learning in educational contexts (Brame 2016). The organisation may choose videos *because* of their (presumed) helpfulness in revealing reasons to users; in that case, their attempt at transparency is clearly not manipulative. But if the organisation opts for videos because they reckon that it will win them favours with users and scholars interested in algorithmic transparency, their influence is manipulative. It is *not* explained by the aim to reveal reasons to users, but tries to achieve some other end effectively.

---

[19] Thanks to an anonymous referee for suggesting the last point.

In short, algorithmic transparency has manipulative potential because the providers of said transparency may be transparent in ways that are simply indifferent to informational quality and revealing pertinent reasons to the users. Instead, they may be much more interested in inducing certain behaviours, such as continued or increased use of and reliance on the system that the algorithm operates in.[20]

---

[20] An important set of question concerns the motives that determine whether or not the attempt at algorithmic transparency was manipulative. First, whose motives count? The 'providers' of algorithmic transparency, like the FICO, are often corporations or other institutions, and there is a large debate about whether or not to think of them as group agents, or mere collectives of individuals (List and Pettit 2011). So far, accounts of manipulation rely on a notion of intention that is at least contentious to ascribe to such groups or artificial entities. Since the ultimate criterion for manipulation on the indifference view is an *explanation* of an influence, it is at least possible to give such an explanation independently of intention but instead in terms of function or purpose, which may more easily be ascribed to groups and artificial agents cf. Klenk (2022). Related to that question is the question of how to determine which amongst the many of motives that reside within an individual agent (or are 'distributed' across collectives of individuals) count toward the assessment of manipulation. For example, a manager may, next to the aim to reveal reasons to their employee, be interested in fulfilling their duty, finishing work that day, and so on. More pertinently, Barclay and Abramson (2021) demonstrate that there are many roles and motives that may legitimately be associated with a given algorithmic system. A tentative suggestion on behalf of the indifference view is that the motive to reveal reasons need not be the only or primary motive (which seems overly demanding) but at least a *causal source* for chosen means of influence, i.e. the chosen influence would be chosen across a range of counterfactual contexts (Lagnado et al. 2013). This would account for to the intuition that manipulative influences are such that the manipulator all too easily forgoes the aim to reveal reasons (which may be present) in favour of the aim to be effective. Tentative as this suggestion is, it would have some bearing on the practical question of how to regulate manipulative algorithmic transparency. For instance, regulation should aim to encourage robust motives to reveal reasons. Their presence

Applied to Wang's example of the FICO algorithm, the indifference view suggests the following picture. If transparency about the FICO algorithm is manipulative, the manipulativeness does *not* lie in the process in which users process the provided information (does it exploit *vulnerabilities*?) or the effects the transparency has on the user (does it *exploit* users?). Instead, the manipulativeness lies in the purpose of the transparency. In the non-manipulative case, the aim is to contribute to the user's deliberation. Inquiry is an important case of deliberation. Credit scoring systems raise a heap of important questions for deliberation. For instance, 'Insofar as I want a credit, how should I behave to get a good credit score?' The FICO transparency is, conceivably, a good-faith contribution to this inquiry. Another question is 'How does this system actually make decision?.' Again, transparency in the FICO case may actually be intended as a contribution to that question. However, there is surely manipulative potential in the FICO case, tool. The purpose of transparency about the FICO case may not be to genuinely contribute to any question that users may have. Instead, the aim may simply be to effectively generate a certain belief such as 'the FICO algorithm is good' or a certain type of behaviour. In that case, the system is clearly manipulative, as Wang warned, and the indifference view tells us why.

Interestingly, Wang offers several remarks that are well aligned with the indifference view already and I want to suggest that some of his observations can fruitfully be understood in light of the indifference view of manipulation.

---

could be assessed by assessing which of the available means of influence – some more, some less reason-revealing, were, in fact, chosen by the influencer. Ultimately, however, this does not fully answer the question of *whose* motives count, and the tentative suggestion would need to be developed further. I thank an anonymous referee for pressing this point.

First, Wang's (2022, 2023) emphatic emphasis on power disparities in commercial or political contexts serves as a useful reminder that algorithmic transparency in the informational sense may seem to be beneficial to stakeholders (e.g. users, and regulators) while the true aims of the deployer of the algorithm may be quite different from serving stakeholders' interests. Deployers of algorithms may thus be disingenuous about their true motives for algorithmic transparency in three relevant ways. They may intend to mislead stakeholders through algorithmic transparency, they may dissimulate their reasons for pushing for algorithmic transparency, or – and this is what the indifference view emphasises – they may use methods for algorithmic transparency for the wrong kinds of reasons. As Wang puts it in the case of the FICO algorithm, their pursuit of algorithmic transparency "does not mean that the FICO Score really *cares* about credit users' true interests" (Wang 2022, p. 19, emphasis added). From the perspective of the indifference view, the provider of the FICO algorithm may manipulate because they go about their transparency with some motive other than to actually reveal reasons to the user. This element of indifference or carelessness in the possible motives behind algorithmic transparency is aptly observed and nicely links up with the indifference account of manipulation.

Second, Wang also comments on the moral problem associated with the manipulation that results from algorithmic transparency in ways that are not readily compatible with vulnerability view, but valuable if seen in the light of the indifference view. Wang (2023) aptly observes that a problem with manipulation is epistemic, and collective insofar as it hampers our collective ability to deliberate and that it requires a collective, political solution. In particular, he suggests that the real model is political and that we have a duty to support collective deliberation. As he puts it, "we as society have the duty to build algorithmic systems that can ensure the healthy development of humans' *deliberative capacity*" (Wang 2023, p. 6). These remarks reflect one of the core insights of the indifference view, namely that manipulation somehow hinders or at least does not reliably promote deliberation.

If seen in this light, some of the features that Wang (2022) explains as manipulation based on the vulnerability view – such as norm-objectification – turn out to be consequences of manipulative influence on the indifference view. For example, suppose the operators of the FICO algorithm chose their informational influences not based on their propensity to reveal reasons to their users, but based on whether they will make the users like or endorse the FICO algorithm. In that case, it is tempting to think that that may make users fail to consider other possible alternatives to set up credit systems. Again, from the perspective of the indifference view, Wang has aptly described a possible result of manipulation.[21].

Algorithmic transparency thus may or may not lead to norm-objectification, and we can leave open whether any step of that process involves exploited vulnerabilities. It is still possible, and perhaps likely, that the deployer of an algorithm may be careless or indifferent in deploying the means of transparency. On the one hand, it will surely be explained by the aim to do something in the vain of transparency. But, on the other hand, the particular method of achieving transparency – a video, a text, and so on – may not be designed such as to be genuinely informative and reveal reasons to users. In that situation, the algorithmic influence may qualify as manipulation in the sense of being careless or indifferent influence.

---

[21] More broadly, and beyond the credit system that Wang discusses, the practice of consciousness raising, cf. Keane (2016), can be interpreted as a way to come to question fixed social structures and – insofar as these structures are to an extent malleable and constructed – it would be a mistake to consider them fixed. The indifference view may – even on a narrow reading, and as a purely contingent, empirical matter, explain how the very process of consciousness raising does not get off the ground as a result of manipulative transparency, insofar as influence that is indifferent reason-revealing *may* (contingently) end up being not reason-revealing influence. It is important to emphasise, again, that this is an empirical question. I am not aware that it has, in specific detail, been explored yet. There is, however, relevant anecdotal evidence from education or training which, in many areas, starts out being geared toward effective influence (simply getting the student to perform a task) and then more and more toward understanding (getting the student to understand why and how the task is performed).

Therefore, the indifference view secures the conclusion about the manipulative potential of algorithmic transparency. There is a sense of manipulation that not only resonates with Wang's general remarks and apt observations about the social- and power-related implications of algorithmic transparency but also save his argument. The influence that results from algorithmic transparency may be indifferent or careless and, therefore, constitute manipulation.

The indifference view also provides a fruitful lens to explore the manipulative potential of algorithmic transparency further. For instance, the indifference view suggests that we should think carefully about what kind of *means* of achieving transparency are best at revealing reasons to users.[22] Mere *transparency* may not suffice. For instance, as Lorenz-Spreen et al. (2021) point out in a different albeit related context, merely making transparent to users that they are now seeing a personalised ad does not significantly alter their decisions. Arguably, more than simple informational transparency is required to reveal reasons to users. So, not being indifferent to users' deliberation may thus mean that one must engage significantly with users' perspectives to get the message across. These explorations could draw on concrete empirical explorations of (requirements for) algorithmic transparency that already exist. For example, a study by Dexe et al. (2020) exemplifies how the value-sensitive design approach can be used to explore transparency while drawing on the contributions from stakeholders. Building on the indifference view, future work from such a design perspective along these lines could explicitly address what it would take to reveal reasons to relevant users, which could serve, for instance, as a guideline for the 'providers' of algorithmic transparency.[23]

---

[22] Though only facts about the manipulator matter for the definition of manipulation (see section 4.4), some of those facts will be facts about what manipulators believe or assume *about their targets* insofar as what it means to *reveal* reasons to someone is at least partly determined by that person's psychology. As discussed above, it is still facts *about the manipulator* (their beliefs, etc.) that matters for determining whether something is manipulation. But insofar as we strive for non-manipulation in our interactions, or aim for design for non-manipulative transparency, we need to form a conception of what it means to reveal reasons to users. Hence, non-manipulators need to form a perception of people's vulnerabilities in order to determine what it means to reveal reasons to them. I thank an anonymous referee for prompting me to clarify this point.

There are also notable open questions about the manipulative potential of algorithmic transparency from the indifference view's perspective that, as pointed out in the introduction, cannot be answered here. Next to future work on refining and explicating the indifference view, which touches on rather philosophical questions about the nature and ethics of manipulation and the underlying ideal of deliberation, in particular concerning the question of how to assess the aims of influencers for manipulation, there are difficult and important questions about operationalising guidelines for non-manipulative transparency, reliable methods to detect manipulative transparency 'in the field,' and investigations of the effects of manipulative transparency. Some tentative suggestions in these directions are to explore value sensitive design approaches (Friedman and Hendry 2019) under the heading of design for non-manipulative transparency, empirical investigations into the motives of providers of transparency, as well as modelling approaches to study the effects of manipulative transparency. If the indifference view inspires further exploration of the manipulative potential of algorithmic transparency, along those lines or others, then this article has reached its goal.

---

[23] Thanks to an anonymous referee for pointing me in the direction of research that already addresses these questions from a design perspective.

## 5 Conclusion

Algorithmic transparency is often regarded as an unequivocally good goal in scholarly and regulatory debates about the societal implications of algorithms. These debates are enriched by a critical perspective that suggests that algorithmic transparency may harbour manipulative potential. So far, however, that perspective rested on a shaky view of manipulation, the vulnerability view.

Therefore, I suggested here an improved notion of manipulation, the indifference view, that salvaged Wang's key insight about hitherto under-acknowledged manipulative aspects of algorithmic transparency. The indifference view suggests that manipulation is a purposeful influence that is not explained by the aim to reveal reasons to the interlocutor. The algorithmic transparency providers may often choose means to achieve transparency for motives other than revealing reasons to users – they may, for example, be interested in leaving a certain impression with users, or complying with regulatory demands. Insofar as these motives replace or crowd out the motive to genuinely contribute to people's understanding of the algorithm, there will indeed be manipulative transparency.

Thus, Wang – and other proponents of the critical perspective on algorithmic transparency – are on the right track, and the indifference view explains why.

Future investigations should explore open questions about the indifference view itself, and assess algorithmic transparency in general or in concrete cases for signs of manipulation, which will plausibly require both philosophical as well as empirical approaches. A notable question that has not been discussed concerns the ethical dimension: future discussions should also focus on the wrong-making features of manipulative transparency, as understood by the indifference view.[24]

[24] I thank the team at the Delft Digital Ethics Centre and two very constructive, meticulous, and helpful anonymous referees for valuable feedback on an earlier version of this paper.

# 6   References

Ananny, M., & Crawford, K. (2018). Seeing without knowing: Limitations of the transparency ideal and its application to algorithmic accountability. *New Media and Society, 20*(3), 973–989.

Bannister, F., & Connolly, R. (2011). The Trouble with Transparency: A Critical Review of Openness in e-Government. *Policy & Internet, 3*, 158–187. doi:10.2202/1944-2866.1076.

Barclay, I., & Abramson, W. (2021). *Identifying Roles, Requirements and Responsibilities in Trustworthy AI Systems*. https://arxiv.org/pdf/2106.08258.

Barnhill, A. (2014). What is manipulation? In C. Coons & M. Weber (Eds.), *Manipulation*: *Theory and practice* (pp. 51–72). Oxford: Oxford University Press.

Baron, M. (2014). The mens rea and moral status of manipulation. In C. Coons & M. Weber (Eds.), *Manipulation*: *Theory and practice* (pp. 98–109). Oxford: Oxford University Press.

Bicchieri, C. (2006). *The grammar of society: The nature and dynamics of social norms*. Cambridge: Cambridge University Press.

Bowles, S., & Gintis, H. (2013). *Cooperative species: Human reciprocity and its evolution*. Princeton, N.J.: Princeton University Press.

Brame, C. J. (2016). Effective Educational Videos: Principles and Guidelines for Maximizing Student Learning from Video Content. *CBE life sciences education.* doi:10.1187/cbe.16-03-0125.

Dexe, J., Franke, U., Nöu, A. A., & Rad, A. (2020). Towards Increased Transparency with Value Sensitive Design. In *International Conference on*

*Human-Computer Interaction* (pp. 3–15). Springer, Cham. doi:10.1007/978-3-030-50334-5_1.

Elster, J. (2015). Social norms. In J. Elster (Ed.), *Explaining Social Behavior*: *More Nuts and Bolts for the Social Sciences* (pp. 347–364). Cambridge University Press.

Estop, J. D. S. (2014). WikiLeaks: From Abbé Barruel to Jeremy Bentham and Beyond. *Cultural Studies ↔ Critical Methodologies, 14*, 40–49. doi:10.1177/1532708613507888.

Felzmann, H., Fosch-Villaronga, E., Lutz, C., & Tamò-Larrieux, A. (2020). Towards Transparency by Design for Artificial Intelligence. *Science and engineering ethics, 26*, 3333–3361. doi:10.1007/s11948-020-00276-4.

Franke, U. (2022). How Much Should You Care About Algorithmic Transparency as Manipulation? *Philosophy & Technology, 35*, 1–7. doi:10.1007/s13347-022-00586-4.

Friedman, B., & Hendry, D. (2019). *Value sensitive design: Shaping technology with moral imagination*. Cambridge, Massachusetts: The MIT Press.

Gorin, M. (2014a). Do Manipulators Always Threaten Rationality? *American Philosophical Quarterly, 51*(1), 51–61.

Gorin, M. (2014b). Towards a theory of interpersonal manipulation. In C. Coons & M. Weber (Eds.), *Manipulation*: *Theory and practice* (pp. 73–97). Oxford: Oxford University Press.

Hanna, J. (2015). Libertarian Paternalism, Manipulation, and the Shaping of Preferences. *Social Theory and Practice, 41*(4), 618–643.

Hurley, M., & Adebayo, J. (2016). Credit scoring in the age of big data. *Yale J.L & Technology, 18*(148), 148–216.

Keane, W. (2016). *Ethical life: Its natural and social histories*. Princeton: Princeton University Press.

Klenk, M. (2020). Digital Well-Being and Manipulation Online. In C. Burr & L. Floridi (Eds.), *Ethics of Digital Well-Being*: *A Multidisciplinary Perspective* (pp. 81–100). Cham: Springer.

Klenk, M. (2021a). Interpersonal Manipulation. *SSRN Electronic Journal*. doi:10.2139/ssrn.3859178.

Klenk, M. (2021b). Manipulation (Online): Sometimes Hidden, Always Careless. *Review of Social Economy*. doi:10.1080/00346764.2021.1894350.

Klenk, M. (2022). Manipulation, injustice, and technology. In M. Klenk & F. Jongepier (Eds.), *The Philosophy of Online Manipulation* (pp. 108–131). New York, NY: Routledge.

Klenk, M., & Jongepier, F. (2022). Manipulation Online: Charting the field. In M. Klenk & F. Jongepier (Eds.), *The Philosophy of Online Manipulation* (pp. 15–48). New York, NY: Routledge.

Kossow, N., Windwehr, S., & Jenkins, M. (2021). *Algorithmic transparency and accountability*. https://knowledgehub.transparency.org/assets/uploads/kproducts/Algorithmic-Transparency_2021.pdf.

Lagnado, D. A., Gerstenberg, T., & Zultan, R. (2013). Causal responsibility and counterfactuals. *Cognitive science, 37*, 1036–1073. doi:10.1111/cogs.12054.

List, C., & Pettit, P. (2011). *Group Agency: The possibility, design, and status of corporate agents*. Oxford: Oxford University Press.

Lorenz-Spreen, P., Geers, M., Pachur, T., Hertwig, R., Lewandowsky, S., & Herzog, S. M. (2021). Boosting people's ability to detect microtargeted advertising. *Scientific Reports, 11*, 15541. doi:10.1038/s41598-021-94796-z.

Mills, C. (1995). Politics and Manipulation. *Social Theory and Practice, 21*(1), 97–112.

Noggle, R. (1996). Manipulative Actions: A Conceptual and Moral Analysis. *American Philosophical Quarterly, 33*(1), 43–55.

Noggle, R. (2018). The Ethics of Manipulation. In E. N. Zalta (Ed.), *Stanford Encyclopedia of Philosophy*: *Summer 2018* (2018th ed.).

Noggle, R. (2020). Pressure, Trickery, and a unified account of manipulation. *American Philosophical Quarterly, 57*, 241–252. doi:10.2307/48574436.

Susser, D., Roessler, B., & Nissenbaum, H. (2019). Technology, autonomy, and manipulation. *Internet Policy Review, 8*, 1–22. doi:10.14763/2019.2.1410.

Susser, D., Roessler, B., & Nissenbaum, H. F. (2018). Online Manipulation: Hidden Influences in a Digital World. *Georgetown Law Technological Review, 4*, 1–45. doi:10.2139/ssrn.3306006.

Wachter, S., Mittelstadt, B., & Russell, C. (2018). Counterfactual Explanations without Opening the Black Box: Automated Decisions and the GDPR. *Harvard Journal of Law & Technology, 31*, 821–887. doi:2018.

Wang, H. (2022). Transparency as Manipulation? Uncovering the Disciplinary Power of Algorithmic Transparency. *Philosophy & Technology, 35*, 1–25. doi:10.1007/s13347-022-00564-w.

Wang, H. (2023). Why Should We Care About the Manipulative Power of Algorithmic Transparency? *Philosophy & Technology, 36*, 1–6. doi:10.1007/s13347-023-00610-1.

Whitehead, A. N. (1911). *An introduction to mathematics*. E-book by Project Gutenberg (originally Williams & Norgate, London).

Winfield, A. F., Booth, S., Dennis, L., Egawa, T., Hastie, H., Jacobs, N., et al. (2021). IEEE P7001: A Proposed Standard on Transparency. *Frontiers in Robotics and AI, 8*.